\documentclass[conference]{IEEEtran}
\IEEEoverridecommandlockouts

\usepackage{cite}

\usepackage{amsmath,amssymb,amsfonts}
\usepackage{algorithmic}
\usepackage{graphicx}
\usepackage{textcomp}
\usepackage{xcolor}
\usepackage{booktabs}
\usepackage{array}
\usepackage{url}
\usepackage{tikz}
\usetikzlibrary{arrows.meta,positioning,shapes.geometric,calc}
\usepackage[hidelinks]{hyperref}
\usepackage[protrusion=true,expansion=true]{microtype}

\selectfont

\begin{document}

\title{\textbf{\textit{When Big Data Becomes a Curse:}}\\Spatial Heterogeneity and the Limits of\\Learning from Passive Acoustic Monitoring Data}

\author{
    \IEEEauthorblockN{Gabriel Spadon$^{\star}$ \thanks{$\star$~Corresponding author.}}
    \IEEEauthorblockA{
        \textit{Faculty of Computer Science} \\
        \textit{Dalhousie University} \\
        Halifax, NS, Canada \\
        spadon@dal.ca
    }
    \and
    \IEEEauthorblockN{Wayne Renaud}
    \IEEEauthorblockA{
        \textit{Faculty of Computer Science} \\
        \textit{Dalhousie University} \\
        Halifax, NS, Canada \\
        wayne.renaud@dal.ca
    }
    \and
    \IEEEauthorblockN{Priyanka Aravindan}
    \IEEEauthorblockA{
        \textit{Faculty of Computer Science} \\
        \textit{Dalhousie University} \\
        Halifax, NS, Canada \\
        priyanka.aravindan@dal.ca
    }
}

\maketitle
\bstctlcite{IEEEtran:BSTcontrol}

\begin{abstract}
Passive Acoustic Monitoring produces large archives whose recordings are clustered by deployment, season, station identifier, and acquisition configuration. We analyze 908{,}072 AIS-labeled 679-second recordings from 38 deployments, 20 Atlantic Canadian station identifiers, and 21 receiver positions. The AIS-contact prior varies by more than 200-fold, and per-deployment screening distributions require local interpretation. Bidirectional cross-season transfer over the 18 station identifiers observed in both seasons predicts station identity above the 5.56\% uniform-chance level, with balanced accuracy of 15.9\% for AIS-contact and 16.4\% for no-AIS-contact recordings. The same descriptors predict the two hydrophone models at 74.1\% and 85.6\% balanced accuracy, respectively, but hydrophone model is strongly confounded with season and other deployment-level acquisition differences. On a retrospectively screened and capped benchmark of 54{,}507 recordings, repeated station-grouped holdout yields an ROC-AUC of 0.612 with a station-bootstrap 95\% interval of 0.582 to 0.648, compared with 0.661 under a random-window diagnostic. Their paired difference is 0.049 (0.040 to 0.056). Removing raw energy changes unseen-station ROC-AUC from 0.612 to 0.603, while an exploratory training-station scale analysis is non-monotonic. These results show that random-window validation overstates transfer to unseen station identifiers in this corpus. They support dependence-aware validation and broader independent spatial sampling, while spatial-expert models remain a hypothesis rather than an established remedy.
\end{abstract}


\section{Introduction}
The ocean is a dynamic acoustic environment, and recent technological advances allow long-duration, duty-cycled monitoring. The ability to capture these sounds is significant because it provides insight into issues such as the strike risk faced by endangered whales~\cite{Blondin2025:VesselStrikeRisk} and anthropogenic noise driven by ship traffic~\cite{Erbe2019:ShipNoiseReview}. Autonomous bottom recorders and cabled observatories can collect broadband underwater sound over extended periods, resulting in archives that exhibit defining characteristics of Big Data, {\it i.e.}, large volume, rapid accumulation, and diverse content. The field has consequently adopted machine learning, often with the expectation that greater data volume will produce more reliable detectors of vessels, marine mammals, and changes in the acoustic environment. Drawing on an AIS-labeled, multi-station corpus from Atlantic Canada, we show that naive pooling can combine strongly heterogeneous and dependent clusters, thereby overstating the independent support available for transfer.

Underwater sound is not merely an abstract feature vector such as those derived from tabular data; it is a physical field in which a source is filtered by range, depth, the sound-speed structure of the water column, and reflections from the seabed and surface before reaching a sensor~\cite{Urick1983:UnderwaterSound}. Passive Acoustic Monitoring (PAM) samples this field at fixed receiver positions, so each deployment observes a distinct received-field distribution, propagation regime, and local vessel-traffic context. Pooled analyses can treat accumulated observations as a single population even though the observations remain tied to their geographic and acquisition origins. Transfer claims should therefore account for spatial structure, a guiding principle of GeoAI~\cite{Janowicz2020:GeoAI}. Spatial heterogeneity produces non-identical distributions across space, while repeated recordings within deployments and station identifiers introduce dependence. Tobler's First Law of Geography states that everything is related to everything else, but near things are more related than distant ones~\cite{Tobler1970:FirstLaw}, encapsulating spatial autocorrelation~\cite{Legendre1993:SpatialAutocorr}.

To illustrate this concept, consider two receivers positioned a few kilometers apart on one continental shelf and a third located a thousand kilometers north in the Labrador Sea. The nearby pair may share storms, passing traffic, and propagation conditions more often than the distant receiver. Tobler's Law can arise through propagation, because source reception depends on the local path, and through sampling, because each receiver observes a local distribution of vessel traffic, bathymetry, and ambient sound. Both mechanisms reduce exchangeability and can make site-agnostic validation unreliable~\cite{Roberts2017:SpatialCV, Ploton2020:SpatialValidation}. Standard supervised-learning guarantees commonly assume that training and deployment pairs are independent and identically distributed ({\it i.i.d.}) under a single law $P(x,y)$. In a multi-station acoustic dataset, however, each observation is indexed by a domain $s$, such as a station or deployment, and may follow its own distribution, {\it i.e.}, $P_{s}(x,y)=P_{s}(y\mid x)P_{s}(x)$.

Spatial heterogeneity means that $P_{s_i}$ and $P_{s_j}$ can differ across domains. Differences in the label marginal $P_s(y)$ produce prior shift, while differences in the feature marginal $P_s(x)$ produce covariate shift. Heterogeneity can also arise between deployments assigned to the same station identifier when placement, orientation, hydrophone model, or local conditions change. A monotone distance-decay relationship formalizes one possible manifestation of Tobler's Law,
\begin{equation}
    \operatorname{Corr}\!\big(z(s_i),z(s_j)\big)=\rho(d_{ij}),\qquad \rho'(d)\le 0,
    \label{eq:tobler}
\end{equation}
where $d_{ij}$ denotes distance between receiver positions and $z(\cdot)$ represents a recorded quantity. We find no clear global monotonic distance-decay pattern; nearest-neighbor confusion shows only a suggestive excess whose interval includes a uniform-allocation baseline. Accordingly, our broader hypothesis is that \textit{large recording counts can be misleading when volume, velocity, and variety are compounded by spatial heterogeneity, because additional dependent observations may reinforce local context without adding independent spatial support}.

Underwater Acoustic Big Data is usually framed in terms of monitoring, signal processing, or machine learning rather than as a spatial sampling problem. This paper makes three contributions. First, we characterize a 908-thousand-recording corpus while distinguishing station identifiers, physical receiver positions, and station--season deployments. Second, we quantify cross-season station-identifier and hydrophone-model recoverability, explicitly delimiting their acquisition confounding. Third, we evaluate vessel-presence transfer with repeated station-grouped holdout, leave-one-region-out, known-station cross-season transfer, and a random-window diagnostic, resampling station identifiers or deployments for uncertainty.

Throughout the paper, \emph{station} denotes the monitoring program's station identifier, which is also the grouping variable used by the analysis code. The 20 identifiers correspond to 21 receiver positions because station 19 was redeployed approximately 210 km east-southeast between seasons. Thus, station-grouped results are not identical to results indexed by a unique fixed physical location, and this distinction is retained.

The remainder of this paper is organized as follows. Section~\ref{sec:related} reviews prior work on underwater acoustics, Passive Acoustic Monitoring, spatial validation, dataset shift, and domain generalization. Section~\ref{sec:methodology} describes the study region, data provenance, acoustic descriptors, station characterization, and experimental protocols. Section~\ref{sec:results} presents the empirical evidence for spatial and temporal heterogeneity, station and hydrophone-model recoverability, and vessel-detection transfer across validation settings. Section~\ref{sec:discussion} interprets these findings through the perspectives of domain shift, spatial dependence, model validation, and spatially expert learning. Finally, Section~\ref{sec:conclusion} summarizes the principal findings and their implications for learning from spatially distributed acoustic Big Data.

\section{Related Work}
\label{sec:related}

The structure of ocean ambient noise has been characterized since the mid-twentieth century, with canonical spectra distinguishing wind, shipping, and biological sources~\cite{Wenz1962:AmbientNoise}. Anthropogenic noise, primarily from low-frequency shipping, is recognized as a significant stressor for marine life~\cite{Erbe2019:ShipNoiseReview}. Standardized acoustic-habitat metrics and data exchange formats for long-term ambient sound facilitate comparison of measurements across instruments and sites~\cite{Merchant2015:AcousticHabitats, Martin2021:MillidecadeSpectra}. This literature establishes physical and measurement heterogeneity, but it does not directly answer how such heterogeneity affects machine-learning validation and transfer.

\begin{figure*}[!t]
    \centering
    \begin{minipage}[t]{0.49\textwidth}\centering
        \includegraphics[width=\linewidth]{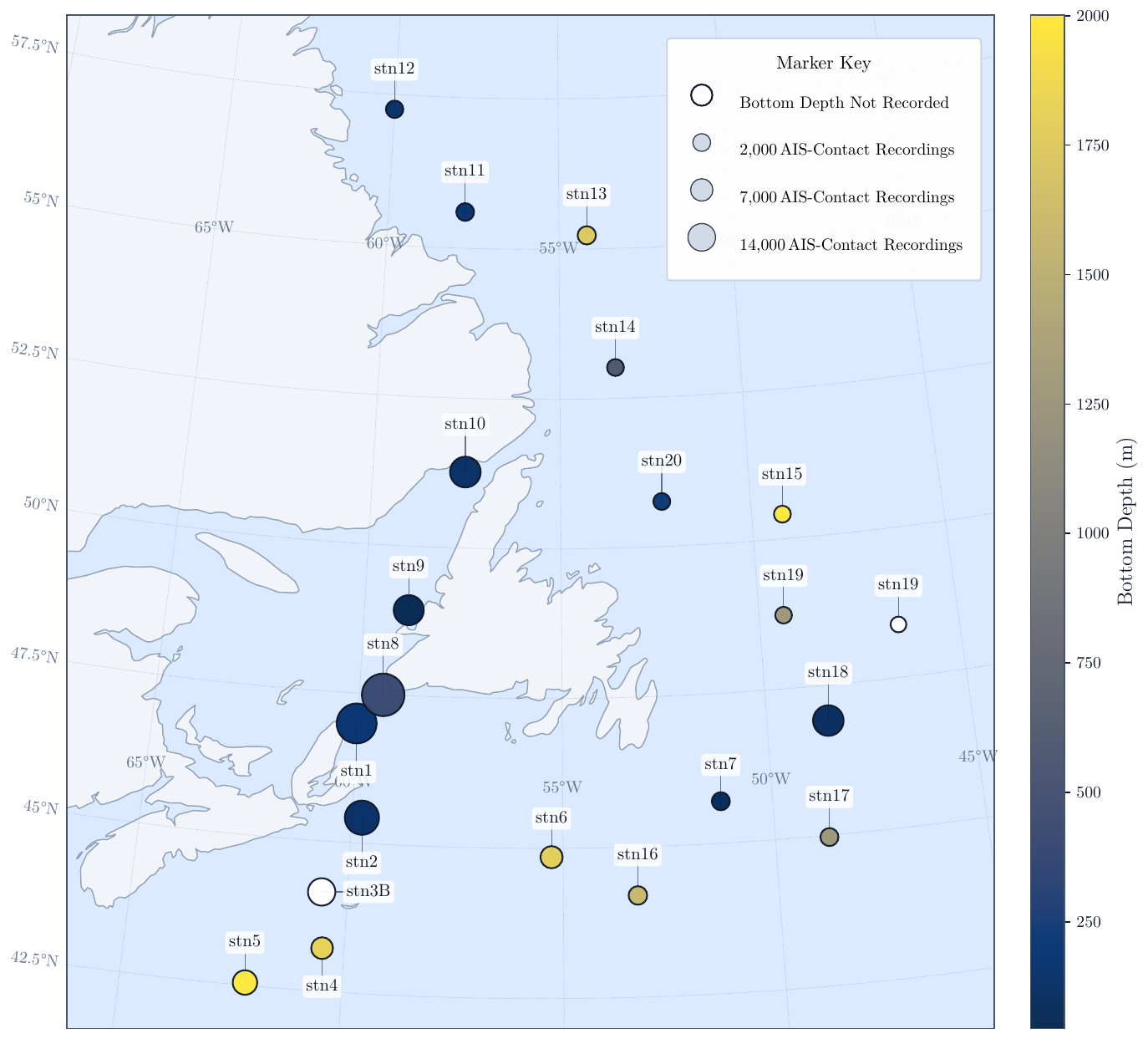}\\
        {\footnotesize (a) deployment positions}
    \end{minipage}\hfill
    \begin{minipage}[t]{0.49\textwidth}\centering
        \includegraphics[width=\linewidth]{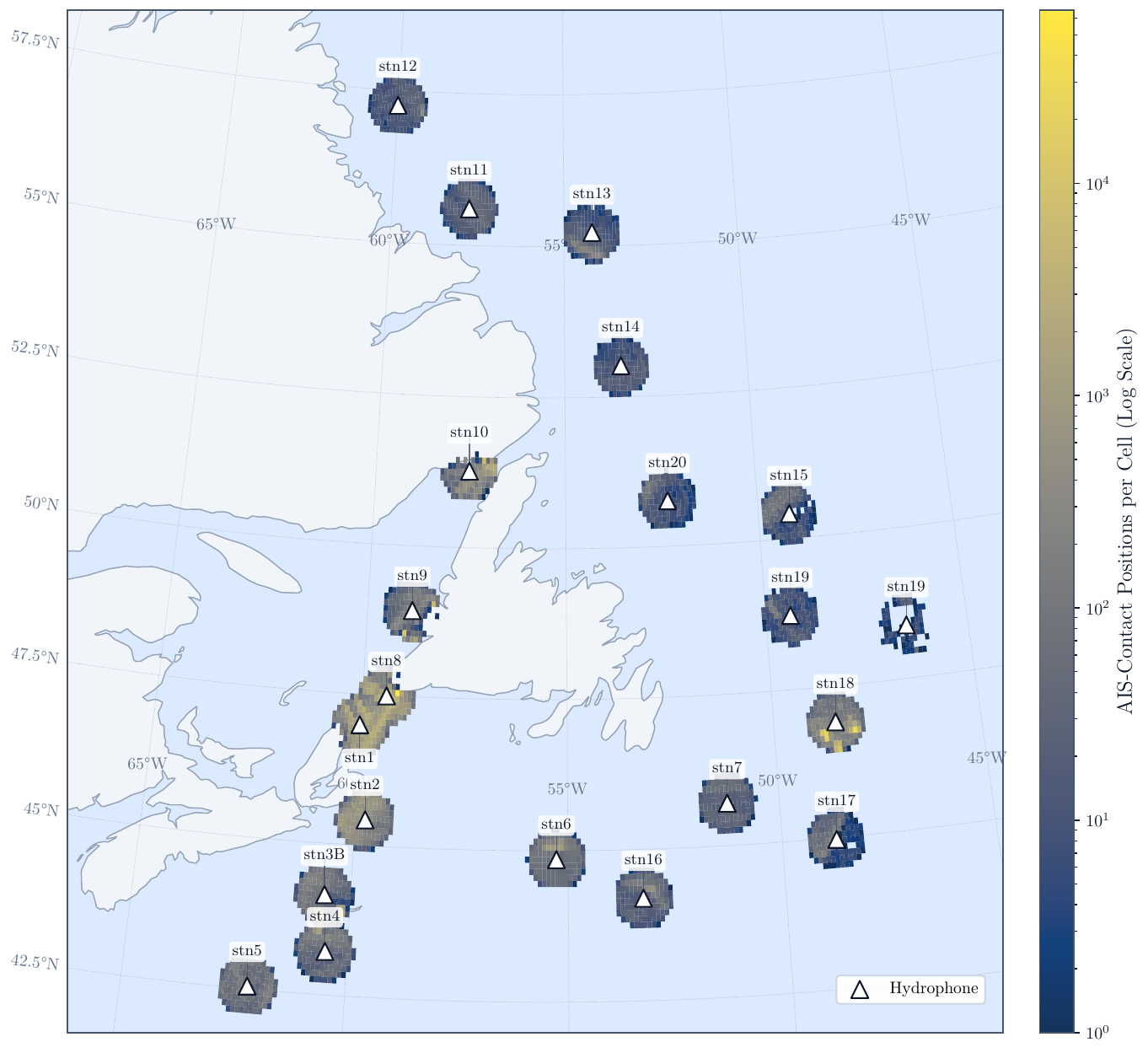}\\
        {\footnotesize (b) AIS encounter density}
    \end{minipage}
    \caption{Array coverage across the Gulf of St. Lawrence and the Northwest Atlantic. (a) Deployment positions are shown, with marker size indicating the number of AIS-contact recordings and color representing bottom depth. Hollow markers correspond to positions without a recorded depth. The array spans approximately 1,600 km and includes bottom depths ranging from 44 to 2,000 m. (b) AIS encounter density under the 50 km matching radius is shown, with closest-approach positions aggregated into 0.1$^{\circ}$ cells on a logarithmic scale.}
    \label{fig:map}
\end{figure*}

Deep neural networks can detect and classify marine bioacoustic signals at scale~\cite{Stowell2022:Bioacoustics, Schall2024:BaleenBenchmark}, and big-data analysis methods for Passive Acoustic Monitoring have been reviewed in the same Atlantic Canada setting considered here~\cite{Kowarski2021:BigDataPAM}. Recent work has addressed cross-site comparison directly by using contrastive representation learning on multi-site marine soundscapes~\cite{Acs2026:ContrastivePAM}. Many evaluations nevertheless pool recordings across deployments or do not isolate whether measured performance reflects the target signal or recurring recording context. The current study addresses this gap through cross-season station attribution and vessel-presence detection at station identifiers absent from training.

The data used by these methods are situated within the broader context of Big Data analytics. In the maritime sector, the Automatic Identification System (AIS) has facilitated large-scale traffic analytics, route mining, and anomaly detection~\cite{Pallotta2013:AISAnomaly}. Recent studies have forecasted long-horizon vessel trajectories and integrated heterogeneous tracking archives at scale~\cite{Spadon2024:VesselForecast, Spadon2024:AISdb}. AIS records provide the operational vessel-proximity labels used in this study. Most maritime Big Data research, however, focuses on vessel-track analysis, while acoustic observations introduce an additional geographic layer mediated by propagation and receiver configuration.

Classical learning theory emphasized the curse of dimensionality~\cite{Bellman1961:Curse}, in which data become sparse as the feature space expands and the sample size required for accurate estimation grows rapidly. Empirical neural scaling laws show that held-out loss can decrease predictably with increases in data, parameters, and computation within the distributions and tasks used to estimate those laws~\cite{Kaplan2020:ScalingLaws}. They do not, by themselves, establish transfer to previously unseen spatial domains. Spatial heterogeneity therefore raises a distinct question: whether added data provide broader domain support or mainly add correlated observations from already represented stations.

A distinct body of literature focuses on this geographic dimension. GeoAI proposes that artificial intelligence applied to geographic data should account explicitly for spatial relationships~\cite{Janowicz2020:GeoAI}, a perspective further supported by advances in data-driven Earth-system science~\cite{Reichstein2019:DLEarthSystem}. A consistent finding is that spatial autocorrelation can inflate estimated model performance when evaluation does not account for spatial structure~\cite{Roberts2017:SpatialCV, Ploton2020:SpatialValidation}. Underwater-acoustic learning presents the same validation concern, although it has received less attention in this literature.

These geographic effects can be expressed as departures from {\it i.i.d.} sampling and as dataset shift, including covariate, prior, and concept shifts~\cite{MorenoTorres2012:ShiftView}. Learning theory addresses error across domains~\cite{BenDavid2010:DomainTheory}, and the domain-generalization literature studies transfer to unseen distributions~\cite{Wang2023:DomainGenSurvey}. Research on shortcut learning~\cite{Geirhos2020:ShortcutLearning} shows that models can exploit incidental correlations, such as acquisition signatures, rather than the intended signal. Camera-trap recognition at unfamiliar locations provides a close geographic analog in which models that perform well on known sites can fail on new ones~\cite{Beery2018:TerraIncognita}. We examine analogous failure modes in underwater acoustics, where propagation, sampling, and acquisition shape features and labels.

Several established approaches address these challenges. Ecology and geosciences have developed spatial, temporal, and hierarchical cross-validation techniques~\cite{Roberts2017:SpatialCV}, shown that random data splits can overstate map accuracy~\cite{Ploton2020:SpatialValidation}, and proposed methods for estimating a model's area of applicability~\cite{Meyer2021:AreaApplicability}. These approaches are well suited to heterogeneous acoustic data but remain uncommon in Passive Acoustic Monitoring validation. This study applies them at the intersection of underwater acoustics, machine learning, and geospatial analysis.

\section{Methodology}
\label{sec:methodology}

\subsection{Study Region and Big Data Profile}
\label{sec:region}
The Gulf of St. Lawrence is a large, semi-enclosed estuarine sea. The St. Lawrence River discharges freshwater into the Gulf, which connects to the open Atlantic through the Cabot Strait between Cape Breton and southwestern Newfoundland and through the Strait of Belle Isle to the north. The deep Laurentian Channel facilitates exchange between the open shelf and the Gulf interior. Strong stratification, a persistent summer cold intermediate layer, and seasonal ice contribute to sound-speed and propagation conditions that can differ from those on the open shelf and slope~\cite{Delarue2018:ESRFReport, Urick1983:UnderwaterSound}.

\begin{table}[!b]
    \caption{The acoustic corpus as a Big Data problem.}
    \label{tab:bigdata}
    \centering
    \renewcommand{\arraystretch}{1.1}
    \begin{tabular}{@{}p{1.5cm}p{6.0cm}@{}}
        \toprule
        \textbf{Dimension} & \textbf{Evidence in this corpus} \\
        \midrule
        \textit{Volume} & Approximately 42 TB in the source acoustic archive reduced to 908{,}072 exact-count recordings; 36{,}347{,}729 canonical window--ping join rows, including 35{,}654{,}802 matched AIS-ping rows from the buffered 50 km join. \\
        \textit{Velocity} & Nominal 679-second recordings on a 1,200-second duty cycle across two seasons (2015/16 and 2016/17); monthly AIS archives; deployment distributions shift between seasons. \\
        \textit{Variety} & 20 station identifiers, 21 receiver positions, 4 regions, 2 hydrophone models, bottom depths up to 2 km, and 0 to 10 strict-window vessels per classified recording. \\
        \textit{Veracity} & Per-deployment \emph{good}, \emph{nominal}, and \emph{outlier} screening tiers; vessel-band-contrast medians of 15 to 51 dB; AIS proximity does not establish acoustic audibility. \\
        \textbf{\textit{Spatial}} & Recordings cluster within 38 deployments and 20 station identifiers; cross-season station attribution is weakly above chance, acquisition configuration is readily recoverable, and station-grouped vessel detection loses skill relative to random-window validation. \\
        \bottomrule
    \end{tabular}
\end{table}

The monitoring array in Fig.~\ref{fig:map}(a) has receiver positions both within the Gulf and throughout the surrounding Northwest Atlantic. Deployments are located near the Cabot Strait and along the west Newfoundland coast, across the Scotian Shelf and Slope south of Nova Scotia, over the Grand Banks and Flemish Cap east of Newfoundland, and along the Labrador Sea and shelf to the north. The array covers approximately $15^{\circ}$ of latitude ($42.5^{\circ}$N to $57.3^{\circ}$N, roughly 1{,}600 km) and bottom depths from 44 m to about 2{,}000 m. The 20 station identifiers map to 21 receiver positions because station 19 moved from $(48.729^{\circ}\mathrm{N},49.381^{\circ}\mathrm{W})$ in 2015/16 to $(48.380^{\circ}\mathrm{N},46.525^{\circ}\mathrm{W})$ in 2016/17, approximately 210 km east-southeast.

The source program used AMAR G3R4 recorders with HTI-99-HF or M36-V35-100 hydrophones and recorded 8 kHz and 250 kHz acoustic channels~\cite{Delarue2018:ESRFReport, NOAA2025:ESRFRawAudio}. The present analysis uses only the 8 kHz channel. Each low-band FLAC contains 24-bit mono audio with a nominal duration of 679 seconds and constitutes one analysis window; the recordings are not subdivided into shorter labeled windows. The source archive is approximately 42 TB, and the exact-count corpus contains 908{,}072 such recordings. Vessel labels are generated through a spatial and temporal join with Canadian Coast Guard AIS reports, producing 36{,}347{,}729 canonical window--ping rows, of which 35{,}654{,}802 are matched AIS-ping rows under the buffered 50 km branch. These join-row counts are provenance records and are not independent acoustic observations.

\begin{table}[!b]
    \caption{Regional characterization of the acoustic dataset.}
    \label{tab:regions}
    \centering
    \renewcommand{\arraystretch}{1.08}
    \setlength{\tabcolsep}{1.0pt}
    \footnotesize
    \begin{tabular}{@{}lccccc@{}}
        \toprule
        \textbf{Region} & \textbf{Stns.} & \textbf{Depl.} & \textbf{Depth (m)} & \textbf{Contact frac.} & \textbf{Contrast (dB)} \\
        \midrule
        Gulf of St.\ Lawrence & 4 & 8 & 44 to 428 & 0.31 to 1.00 & 15 to 42 \\
        Scotian Shelf/Slope & 4 & 7 & 126 to 2002 & 0.10 to 0.62 & 26 to 32 \\
        Grand Banks/Flemish & 6 & 11 & 78 to 1802 & 0.01 to 0.47 & 20 to 43 \\
        Labrador Sea/Shelf & 6 & 12 & 143 to 2000 & 0.005 to 0.07 & 20 to 51 \\
        \bottomrule
    \end{tabular}
\end{table}

\subsection{Labeling and Provenance}
\label{sec:method}
The labeling pipeline first joins AIS reports to recording windows. A report is a candidate match when its great-circle distance to the receiver is at most 50 km and its timestamp lies in the recording interval extended by five minutes before onset and six minutes after offset. The exact-count label is then determined from the number of distinct vessel identifiers whose reports fall inside the unbuffered 679-second recording interval. A \emph{no-AIS-contact} recording has no qualifying report in the buffered join. The AIS-contact classes contain exactly one through ten distinct vessels in the strict interval. The exact-count partition excludes 43{,}704 recordings that have a 50 km match only in the temporal buffer and no vessel in the strict interval; these recordings remain in the separate 951{,}776-recording graded-context table. The resulting exact-count corpus contains 215{,}145 AIS-contact and 692{,}927 no-AIS-contact recordings.

For every vessel contributing to an AIS-contact class, the strict-window closest point of approach is the minimum reported receiver-to-vessel distance during the unbuffered interval. A separate buffer-wide reduction supports the encounter-density map. These quantities describe encounter geometry only and should not be interpreted as measures of acoustic audibility. AIS is also not a complete vessel census, so a no-AIS-contact label denotes the absence of a qualifying reported track, not confirmed acoustic absence of vessels or other sound sources.

Each recording is acoustically summarized after removing the first and last 10 seconds, leaving approximately 659 seconds for feature extraction while label timing continues to use the full 679-second interval. The raw-waveform mean-square energy is retained as \emph{energy mean}. The waveform is then normalized as $(x-\bar{x})/(\operatorname{std}(x)+10^{-12})$, and a short-time Fourier transform with 1{,}024-sample frames and a 512-sample hop yields four amplitude-normalized descriptors: spectral flatness, spectral entropy, peak-to-mean ratio, and frame-energy variance. A Welch power spectral density estimate with 2{,}048-sample segments adds a separate 1{,}025-bin spectrum over 0--4,000 Hz~\cite{Welch1967:PSD}. The classifier uses the five scalar descriptors, not the full Welch spectrum.

For AIS-contact recordings, vessel-band contrast compares mean 5--100 Hz power with the tenth percentile of the same recording's full-band spectrum,
\begin{equation}
    C_{\mathrm{band}}=10\log_{10}\frac{\tfrac{1}{|B|}\sum_{f\in B}\hat{S}(f)}{Q_{10}\{\hat{S}(f)\}},
    \label{eq:contrast}
\end{equation}
where $\hat S$ is the Welch spectrum, $B$ is the vessel band, and $Q_{10}$ is the full-band tenth percentile. This within-recording ratio is invariant to a uniform multiplicative gain, but its denominator is not an independent ambient estimate. It is therefore a spectral-contrast statistic, not a physical SNR, calibrated level, or proof that a matched vessel generated low-frequency energy.

Each recording carries a descriptive screening tier on one of two class-specific scales. For no-AIS-contact recordings, the background score is
\begin{equation}
    v_{\mathrm{bg}}=\mathrm{flatness}+\mathrm{entropy}-\mathrm{peak\mbox{-}to\mbox{-}mean}-\mathrm{energy\ variance},
    \label{eq:bgscore}
\end{equation}
and a spectral anomaly detector log-compresses the 1{,}025-bin PSD, standardizes each bin within a deployment, retains at most 15 principal components, and applies an isolation forest with automatic contamination~\cite{Liu2008:IsolationForest}. AIS-contact recordings use vessel-band contrast without this PSD anomaly detector. For either score $v$, robust statistics are computed separately within deployment and class,
\begin{equation}
    z_i=\frac{v_i-\operatorname{med}(v)}{1.4826\operatorname{MAD}(v)},\quad
    \operatorname{MAD}(v)=\operatorname{med}|v-\operatorname{med}(v)|.
    \label{eq:z}
\end{equation}
When the scaled MAD is zero or non-finite, finite values receive $z_i=0$ and non-finite scores remain non-finite. A recording is \emph{outlier} when $z_i<-3$, its score is non-finite, or, for the no-AIS-contact pool, the PSD detector flags it. It is \emph{good} when its score is at or above the deployment's 75th percentile and it is not an outlier; otherwise it is \emph{nominal}. These tiers rank recordings within deployment and class. They are not calibrated quality measures or indicators of label correctness.

The primary transfer analysis groups by station identifier, not only by station--season deployment. Every deployment assigned to a test station identifier remains outside training, including both physical positions of station 19. Random-window, paired cross-season, and leave-one-region-out evaluations are reported as distinct protocols because they answer different transfer questions. Confidence intervals resample station identifiers for station-grouped analyses and deployments for the random-window diagnostic; individual recordings and fold means are not treated as independent sampling units.

\subsection{Data and Station Characterization}
\label{sec:char}
The 20 station identifiers are organized into four oceanographic regions, as summarized in Table~\ref{tab:regions}. The regions differ in depth, AIS-contact fraction, and vessel-band contrast, and substantial variation also exists within regions. Across the 21 receiver positions, bottom depths range from 44 m to about 2{,}000 m.

\begin{figure}[tb]
    \centering
    \includegraphics[width=\columnwidth]{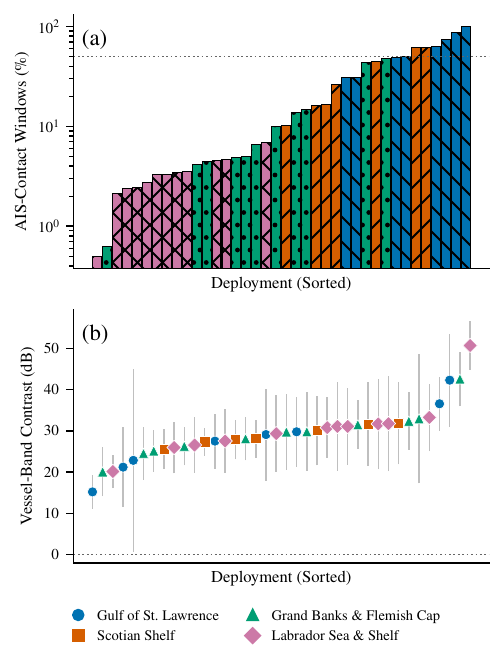}
    \caption{Two observed dimensions of heterogeneity, with one bar or point per deployment and color denoting region. (a) The AIS-contact prior spans more than two orders of magnitude (log axis; dotted line at 50\%). (b) The median vessel-band contrast of AIS-contact recordings ranges from 15 to 51 dB, with substantial within- and between-region variability; whiskers show robust median absolute deviation. The summaries establish that deployment-level distributions are not constant across the corpus.}
    \label{fig:het}
\end{figure}

The primary axis of heterogeneity concerns the decision problem itself. Fig.~\ref{fig:het}(a) presents the 38 deployments sorted by AIS-contact rate. The prior varies by more than $200\times$, ranging from 0.49\% to 99.8\%. This ordering reflects geographic patterns, with Gulf approaches near the Cabot Strait and heavily trafficked Grand Banks stations at the high end and Labrador deployments at the low end. Consequently, a model that internalizes one region's prior can be miscalibrated in another, and a single threshold from pooled data need not be optimal across all regions. This is spatial prior shift.

Heterogeneity is also evident in vessel multiplicity. While most AIS-contact recordings contain a single matched vessel, 73{,}118 contain three or more, and some contain as many as ten. The binary label therefore aggregates geometrically distinct contexts with different numbers, ranges, and emitter bearings. Multiple sources can also overlap with weather, biological, and ambient sound, so the positive class spans acoustically distinct conditions across stations. The label does not establish that every matched vessel contributed audible energy. Figure~\ref{fig:map}(b) shows encounters concentrated along shipping corridors within receiver search footprints, making the positive-label prior location-dependent.

\subsection{Label Screening Analysis}
\label{sec:quality}
Figure~\ref{fig:het}(b) shows that the vessel-band-contrast distribution changes across deployments. This variation can reflect received-field composition, traffic mixture, propagation, hydrophone response, and uncalibrated recording level. The statistic cannot determine whether a matched vessel produced the band energy because distance alone does not determine received level. Negative contrast means only that mean 5--100 Hz power falls below the recording's full-band tenth-percentile reference.

\begin{figure}[t]
    \centering
    \includegraphics[width=\columnwidth]{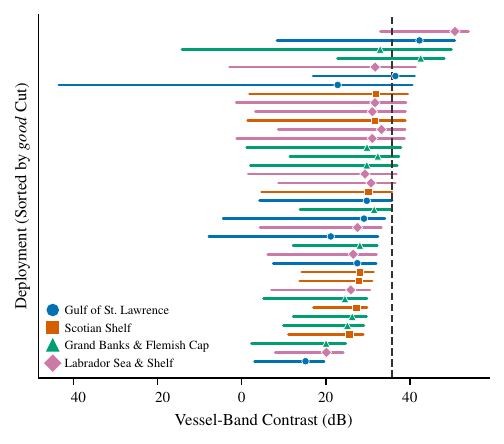}
    \caption{Per-deployment vessel-band-contrast screening boundaries. Each bar spans the local \emph{outlier} to \emph{good} interval, the dot marks the median, and the dashed line is the median \emph{good} cut. The boundaries are descriptive within-deployment ranks.}
    \label{fig:quality}
\end{figure}

Figure~\ref{fig:quality} shows the resulting within-deployment screening boundaries. The \emph{good} threshold ranges from approximately 20 dB to 54 dB, while the \emph{outlier} threshold spans approximately $-43$ dB to $+33$ dB across deployments. Because these thresholds are defined from local percentiles and robust statistics, numerical variation is expected when deployment score scales differ. The figure documents this scale variation; it does not establish that no globally useful operational threshold could be designed. The detector benchmark excludes outliers using these class-specific retrospective tiers, so its performance applies to the retained benchmark, not an unfiltered deployment stream.

\subsection{Experimental Protocol}
\label{sec:protocol}
All classification experiments use the same five scalar descriptors. Vessel-band contrast is excluded because it is defined only for AIS-contact recordings and would reveal the detector label. Unless otherwise stated, the model is a 300-tree random forest with class-balanced weights.

The station-attribution analysis fits separate models to AIS-contact and no-AIS-contact pools after removing outlier and non-finite rows. It uses 198{,}739 AIS-contact and 580{,}711 no-AIS-contact recordings and trains in one season while testing in the other, in both directions, across the 18 station identifiers observed in both seasons. Stations 3B and 7 have no paired-season support and are excluded. Balanced accuracy is the average class recall,
\begin{equation}
    \mathrm{BAcc}=\frac{1}{K}\sum_{k=1}^{K}\frac{\mathrm{TP}_k}{n_k},
    \label{eq:bacc}
\end{equation}
with $K=18$. The chance level is $1/K=0.0556$; the largest class fraction is reported separately only to contextualize plain accuracy. Station-attribution intervals are obtained by bootstrapping the 18 station-specific recalls and averaging them. In parallel, deployment-blocked stratified group cross-validation predicts the two hydrophone models from the same descriptors, with both model classes represented in every fold and station identifiers used as the uncertainty-resampling unit.

Hydrophone-model attribution is not a causal sensor comparison. Among the 18 paired station identifiers, 17 use HTI-99-HF in 2015/16 and M36-V35-100 in 2016/17, while station 2 uses M36-V35-100 in both seasons. Hydrophone model is therefore strongly aligned with season and other deployment-level acquisition changes. Deployment blocking prevents the same deployment from appearing in training and testing, but it does not isolate hydrophone response from season, hardware configuration, or program-wide differences.

The vessel-presence analysis removes non-finite rows and class-specific outliers, then caps each deployment--label cluster at 750 seeded recordings. This produces a 54{,}507-recording benchmark while limiting domination by large deployments. The cap reweights deployments and classes, so F1 and average precision describe this benchmark rather than operational field prevalence. Because outlier exclusion uses the known class, including vessel-band contrast for AIS-contact recordings, it is retrospective benchmark construction rather than deployable label-free preprocessing.

The binding protocol is three repeats of five-fold stratified group cross-validation with station identifier as the group. Every deployment assigned to a held-out station remains unseen during training. This is the study's approximation to transfer to an unrepresented station; station 19's two physical positions are held out together. Three diagnostics answer narrower questions. A random-window split uses repeated stratified fivefold cross-validation without station or deployment grouping. Bidirectional cross-season transfer trains on one season and tests on the other over the same 18 station identifiers, although station 19 changes physical position. Leave-one-region-out evaluates the four observed regions.

A separate scale analysis retains 25, 50, 75, or 100\% of the training station identifiers in each fold while preserving the held-out stations. It uses only the first of the three outer-split repeats, draws one seeded and non-nested training-station subset per fraction and fold, and does not hold the number of training recordings fixed. Its bootstrap resamples test stations and therefore does not include uncertainty from training-subset selection. The analysis is exploratory rather than a controlled learning-curve experiment.

Balanced accuracy and F1 use a probability threshold of 0.5. ROC-AUC and average precision use held-out probabilities; average precision is prevalence-dependent. Marginal intervals use 1{,}000 percentile cluster-bootstrap draws, with station identifier as the resampling unit for grouped protocols and deployment as the unit for the random-window diagnostic. Repeated predictions are summarized within repeat before aggregation. The robustness panel additionally uses default class-balanced histogram gradient boosting and standardized class-balanced logistic regression with the liblinear solver and 2{,}000 iterations. Permutation importance uses ten shuffles per descriptor in each of the 15 unseen-station folds. Its displayed intervals bootstrap folds and are exploratory because folds across repeated partitions are not independent station samples.

\subsection{Paired Protocol Contrast}
Marginal intervals do not test whether the random-window and unseen-station protocols differ because station difficulty affects both estimates. We therefore define
\begin{equation}
    \Delta_{\mathrm{AUC}}=\frac{1}{R}\sum_{r=1}^{R}\left[
    \mathrm{AUC}\!\left(\hat p_{r}^{\mathrm{random}}\right)-
    \mathrm{AUC}\!\left(\hat p_{r}^{\mathrm{unseen}}\right)\right],
    \label{eq:paired-delta}
\end{equation}
where $R=3$ denotes repeated partitions. The two prediction panels contain identical held-out recording identifiers within repeat. Each bootstrap replicate samples station identifiers with replacement and applies the same station multiplicities to both protocols and all repeated runs. This paired resampling preserves covariance induced by shared station difficulty and estimates the protocol contrast directly.

\section{Results}
\label{sec:results}
The marginal class priors, descriptor distributions, and screening boundaries vary across deployments. We first test whether the joint feature distribution retains station-associated structure across seasons.

\begin{table}[t]
    \caption{Station-identifier attribution and hydrophone-model attribution from the five acoustic descriptors.}
    \label{tab:sep}
    \centering
    \renewcommand{\arraystretch}{1.08}
    \setlength{\tabcolsep}{2.5pt}
    \footnotesize
    \begin{tabular}{@{}lcccc@{}}
        \toprule
        \textbf{Pool} & \textbf{BAcc} & \textbf{95\% CI} & \textbf{Chance} & \textbf{Largest class} \\
        \midrule
        \multicolumn{5}{@{}l}{\emph{Predict the station identifier} (18 classes)} \\
        \quad AIS-contact & \textbf{0.159} & 0.072 to 0.252 & 0.056 & 0.214 \\
        \quad no-AIS-contact & \textbf{0.164} & 0.075 to 0.269 & 0.056 & 0.076 \\
        \addlinespace
        \multicolumn{5}{@{}l}{\emph{Predict the hydrophone model} (2 classes)} \\
        \quad AIS-contact & \textbf{0.741} & 0.630 to 0.855 & 0.500 & 0.642 \\
        \quad no-AIS-contact & \textbf{0.856} & 0.797 to 0.903 & 0.500 & 0.517 \\
        \bottomrule
    \end{tabular}
\end{table}

\begin{figure*}[!t]
    \centering
    \includegraphics[width=\textwidth]{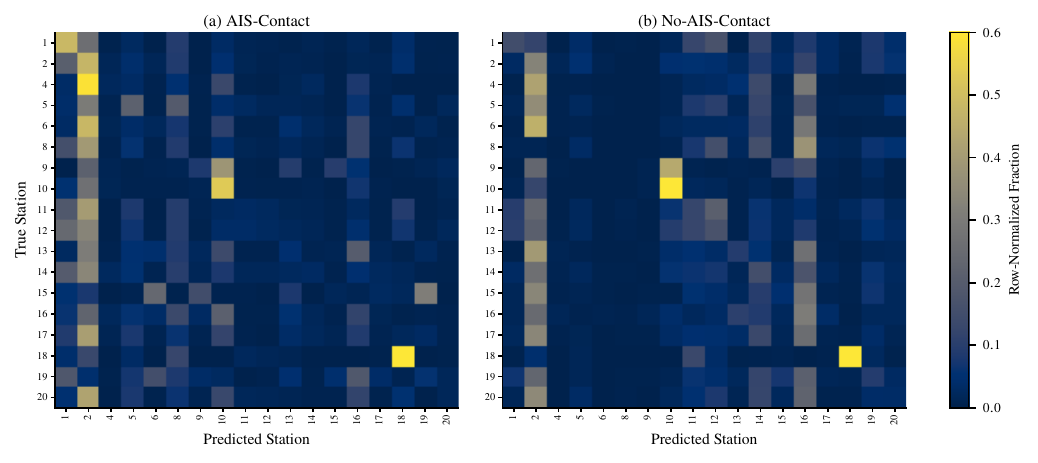}
    \caption{Bidirectional cross-season station-identifier confusion matrices over the 18 paired identifiers, row-normalized, for AIS-contact (a) and no-AIS-contact (b) pools. Every test identifier is represented in opposite-season training. A faint diagonal appears against broad off-diagonal mass, indicating weak station-associated information rather than a clean or strongly transferable station representation.}
    \label{fig:confusion}
\end{figure*}

Station identifier is recoverable above chance, although only weakly. Balanced accuracy reaches 15.9\% for AIS-contact recordings and 16.4\% for no-AIS-contact recordings, compared with the 5.56\% chance level (Table~\ref{tab:sep}). The corrected station-recall bootstrap intervals are 7.2 to 25.2\% and 7.5 to 26.9\%, respectively. Plain accuracy is 29.9\% and 14.4\%, compared with largest-class fractions of 21.4\% and 7.6\%. Because every test identifier has opposite-season examples in training, the experiment measures persistence of a station label rather than classification with unsupported classes. It does not measure a unique fixed physical location for station 19, which moved between seasons. Figure~\ref{fig:confusion} shows broad off-diagonal confusion, so the descriptors hold only a weak station signature.

The same features predict hydrophone model substantially above chance. Deployment-blocked balanced accuracy is 74.1\% in the AIS-contact pool and 85.6\% in the no-AIS-contact pool, against 50\% chance. Relative to their respective chance levels, the point estimates close approximately 48\% and 71\% of the available hydrophone-model margin, compared with approximately 11\% for station attribution. This descriptive comparison is not a direct comparison of task difficulty because the class counts and validation protocols differ. More importantly, the hydrophone-model target is strongly confounded with season and acquisition configuration. The result establishes recoverable acquisition-regime structure, not a causal hydrophone signature or detector shortcut.

We next examine whether station-attribution errors are geographically structured. For each true station identifier $i$, we measure
\begin{equation}
    \kappa_i=\frac{\sum_{j\in\mathcal{N}_3(i)}C_{ij}}{\sum_{j\neq i}C_{ij}},
    \label{eq:kappa}
\end{equation}
where $\kappa_i$ is the share of off-diagonal confusion mass $C_{ij}$ assigned to the three nearest station identifiers, denoted by $\mathcal{N}_3(i)$. With 18 supported identifiers, uniform allocation yields $3/17=0.176$. The observed mean is 0.279 in the AIS-contact pool, with a station bootstrap interval of 0.156 to 0.395, and 0.224 in the no-AIS-contact pool, with an interval of 0.149 to 0.312. Both intervals include the uniform baseline, so the nearest-neighbor excess is suggestive rather than conclusive.

\begin{figure}[t]
    \centering
    \includegraphics[width=\columnwidth]{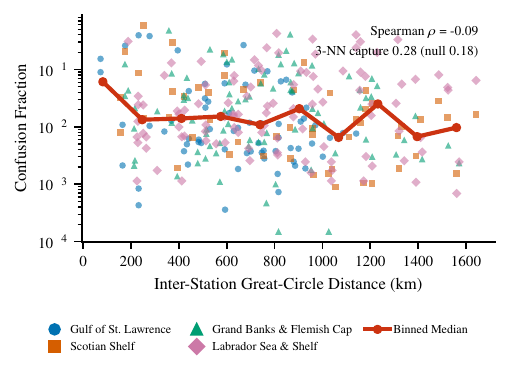}
    \caption{Pairwise station-attribution confusion for AIS-contact recordings versus inter-station great-circle distance, with a binned-median trend. The three nearest identifiers capture 0.279 of off-diagonal mass, compared with 0.176 under uniform allocation, while the descriptive global rank association is weak ($\rho=-0.092$). The distance calculation uses each identifier's first-season coordinate, so station 19's relocation is not represented.}
    \label{fig:decay}
\end{figure}

The direct monotonic association between great-circle distance and pairwise confusion is weak. Spearman's $\rho$ is $-0.092$ for AIS-contact recordings and $0.097$ for no-AIS-contact recordings. The 306 directed pairwise entries share station endpoints and each confusion row is compositional, so ordinary Spearman significance tests are not valid; the coefficients are treated descriptively. Distances use the first-season coordinate associated with each identifier, including station 19. Geographic distance is one component of acoustic similarity, and propagation can make distant receivers similar or nearby ones distinct.

Heterogeneity is both temporal and spatial. Among the 18 station identifiers recorded in both seasons, the median absolute cross-season change is 3.9 percentage points for AIS-contact prior (maximum 26.6) and 1.9 dB for median vessel-band contrast (maximum 21.4 dB); one identifier's contact rate falls from 4.5\% to 0.5\%. These comparisons are between deployments carrying the same identifier, not necessarily identical physical positions in the case of station 19. They show that a station label does not imply a temporally fixed distribution.

Table~\ref{tab:gen} evaluates the vessel-presence detector. The binding unseen-station ROC-AUC is 0.612 with a station-bootstrap 95\% interval of 0.582 to 0.648. The random-window diagnostic reaches 0.661 (0.636 to 0.690). Known-station cross-season transfer gives 0.603 (0.576 to 0.632), and leave-region-out gives 0.601 (0.571 to 0.635). Every ROC-AUC interval lies above 0.5, but the random-window diagnostic gives a higher estimate than station-grouped transfer. The grouped-protocol intervals overlap substantially, so these results do not establish an ordering among known-station cross-season, leave-region-out, and unseen-station transfer.

The paired protocol contrast resolves the random-window versus unseen-station comparison. Random-window validation has a higher AUC, with $\Delta_{\mathrm{AUC}}=0.049$ and a shared-station bootstrap 95\% interval of 0.040 to 0.056. None of the 1{,}000 paired bootstrap replicates is non-positive. Conditional on the observed station mixture and resampling scheme, the advantage is systematic. It documents optimism for this random-window diagnostic without implying that every random split is invalid for every deployment target.

\begin{table}[b]
    \caption{Vessel-presence detection on the capped, retrospectively screened benchmark. AP denotes average precision.}
    \label{tab:gen}
    \centering
    \renewcommand{\arraystretch}{1.1}
    \setlength{\tabcolsep}{2.2pt}
    \scriptsize
    \begin{tabular}{@{}lcccc@{}}
        \toprule
        \textbf{Protocol} & \textbf{BAcc} & \textbf{F1} & \textbf{ROC-AUC} & \textbf{AP} \\
        \midrule
        Random-window diagnostic & 0.615 & 0.604 & 0.661 & 0.664 \\
        Known-station cross-season & 0.576 & 0.564 & 0.603 & 0.583 \\
        Leave-region-out & 0.576 & 0.567 & 0.601 & 0.581 \\
        \textbf{Unseen station} & \textbf{0.585} & \textbf{0.572} & \textbf{0.612} & \textbf{0.591} \\
        \quad gain-invariant features & 0.576 & 0.562 & 0.603 & 0.588 \\
        \bottomrule
    \end{tabular}
\end{table}

\begin{figure}[t]
    \centering
    \includegraphics[width=\columnwidth]{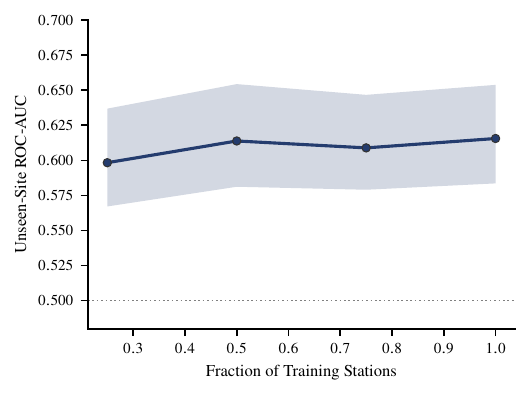}
    \caption{Exploratory unseen-station ROC-AUC against the fraction of available training station identifiers retained in the first outer-split repeat. The non-nested subsets, varying training-set sizes, and overlapping test-station bootstrap intervals do not establish a smooth scaling relationship.}
    \label{fig:scale}
\end{figure}

Training-station coverage has a non-monotonic association with transfer in the exploratory scale experiment (Fig.~\ref{fig:scale}). ROC-AUC values are 0.598, 0.614, 0.609, and 0.616 when 25, 50, 75, and 100\% of available training station identifiers are retained. The intervals overlap throughout. The 100\% value of 0.616 is based on the first outer-split repeat, whereas the primary unseen-station value of 0.612 averages three repeats; the values therefore need not coincide. Because the training subsets are non-nested, training recording counts vary, and training-subset uncertainty is not resampled, the experiment does not support a learning-law claim.

Removing raw energy mean changes unseen-station AUC from 0.612 to 0.603, a point-estimate decrease of 0.009, with strongly overlapping marginal intervals. The other four descriptors are computed after per-recording amplitude normalization. This ablation shows that unseen-station performance is not wholly dependent on raw recording scale. It does not test whether raw scale explains the random-to-unseen gap, which would require the same ablation under both protocols, and it does not identify a causal acquisition effect.

\begin{figure}[t]
    \centering
    \includegraphics[width=\columnwidth]{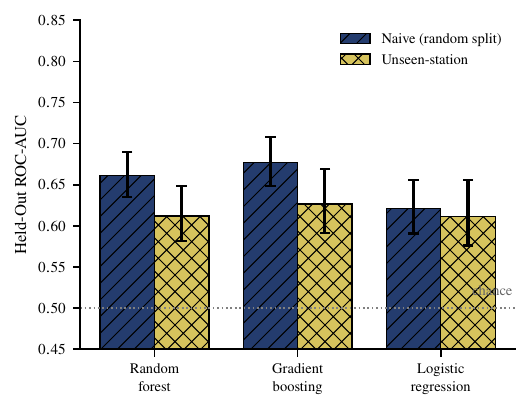}
    \caption{Vessel-presence ROC-AUC under random-window and unseen-station protocols for three classifier families. Bars show held-out estimates and whiskers show cluster-bootstrap 95\% intervals. The descriptive gaps are larger for the two nonlinear models, but no formal model-capacity interaction is tested.}
    \label{fig:robustness}
\end{figure}

\begin{figure}[b]
    \centering
    \includegraphics[width=\columnwidth]{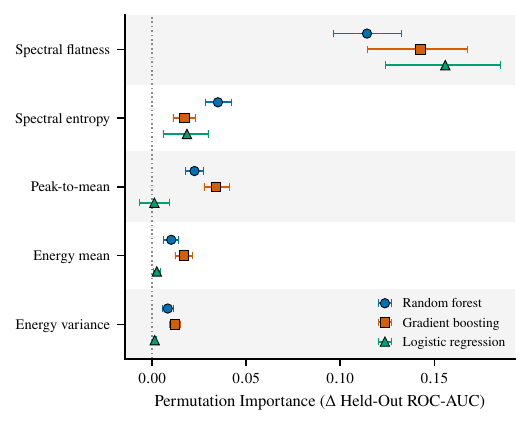}
    \caption{Permutation importance under the unseen-station protocol. Markers show mean decreases in held-out ROC-AUC over ten shuffles per descriptor; whiskers are exploratory fold-bootstrap intervals over 15 folds. Repeated folds are not independent station samples, and importance can be shared among correlated descriptors.}
    \label{fig:importance}
\end{figure}

All three classifier families have higher random-window than unseen-station point estimates (Fig.~\ref{fig:robustness}). Random-window ROC-AUC is 0.661 (0.636 to 0.690) for the random forest, 0.677 (0.649 to 0.708) for histogram gradient boosting, and 0.622 (0.591 to 0.655) for logistic regression. Under unseen-station evaluation, the corresponding values are 0.612 (0.582 to 0.648), 0.627 (0.591 to 0.669), and 0.611 (0.576 to 0.656). The exact gaps are 0.049, 0.051, and 0.010. They appear in both nonlinear models and are small for logistic regression, but no paired interaction test establishes a capacity effect and no mechanism-specific experiment identifies their source. No paired between-classifier comparison is reported, so the classifier ranking remains unresolved.

Among the five descriptors, spectral flatness has the largest mean permutation importance for all three classifier families (Fig.~\ref{fig:importance}): 0.114 for random forest, 0.143 for histogram gradient boosting, and 0.156 for logistic regression. Raw energy mean changes ROC-AUC by 0.010, 0.017, and 0.003, respectively, which is small relative to flatness and consistent with the gain-invariant ablation. Spectral entropy, peak-to-mean ratio, and energy variance have smaller and model-dependent means. Because permutation importance depends on the fitted model, held-out fold, descriptor correlations, and permutation scheme, the result supports predictive reliance on normalized spectral shape among these five descriptors but does not identify a unique or causal acoustic mechanism.

\section{Discussion}
\label{sec:discussion}
The standard supervised pipeline commonly treats held-out samples as exchangeable draws from a common distribution, whereas this corpus has nested dependence: recordings within deployments, deployments within station identifiers, and stations within regions. Prior shift is extreme because AIS-contact rate varies by more than $200\times$ across deployments (Fig.~\ref{fig:het}(a)). Covariate shift is plausible across propagation conditions, ambient fields, hydrophone models, and deployment configurations. If the scientific target is audible vessel sound, an AIS-proximity label can also act as label noise because it denotes reported geometry rather than verified acoustic contribution~\cite{Frenay2014:LabelNoise}. If the target is operational AIS proximity under the stated join rule, the same limitation is part of the target definition. The data do not identify prior, covariate, concept, acquisition, and label effects separately, but grouped evaluation shows that their combined domain structure changes measured transfer~\cite{MorenoTorres2012:ShiftView, BenDavid2010:DomainTheory}.

Under the assumptions of binary domain-adaptation theory, a hypothesis $h$ satisfies
\begin{equation}
    \varepsilon_{T}(h)\le \varepsilon_{S}(h)+\tfrac{1}{2}d_{\mathcal{H}\Delta\mathcal{H}}(\mathcal{D}_S,\mathcal{D}_T)+\lambda,
    \label{eq:bound}
\end{equation}
where $\varepsilon_S$ is source classification error, $d_{\mathcal{H}\Delta\mathcal{H}}$ is a source--target divergence, and $\lambda$ is their best joint error~\cite{BenDavid2010:DomainTheory}. We use this bound only as a qualitative framework because the experiments report ROC-AUC and do not estimate its terms. The higher random-window AUC shows that evaluation-domain construction changes measured discrimination; it is not a direct estimate of source error, divergence, or $\lambda$.

The experiments use five compact descriptors and standard tabular classifiers, not calibrated spectra, waveform models, or deep representations, so they do not establish an upper bound on station attribution or vessel detection. Station attribution covers only 18 paired identifiers, while station-grouped detection covers 20 identifiers from one monitoring program. The identifiers correspond to 21 positions, and the distance analysis represents station 19 by its first-season coordinate. Hydrophone model is strongly confounded with season and acquisition configuration, so no causal instrument comparison is possible. AIS labels establish reported proximity under a fixed 50 km and asymmetric temporal-buffer rule, not audibility or source contribution, and sensitivity to alternative label radii or buffers is not evaluated. The detector benchmark also uses label-dependent retrospective screening and deployment--class capping, so its F1 and average precision do not represent the operational prevalence of an unfiltered stream. Cluster-bootstrap intervals over 18 to 20 station identifiers quantify variation across the observed identifiers but are approximate for broader spatial inference, while leave-one-region-out covers only four regions. The scale analysis adds further design limitations because it uses one repeated partition and does not resample training-subset selection. Finally, the study does not show that adding recordings causes performance to decline. It shows that a large recording count and random-window validation do not guarantee transfer to a new station identifier.

Practical implications depend on the intended deployment. Claims about an unrepresented station require station-grouped validation, whereas deployment-grouped folds answer a narrower question about new deployments from represented station identifiers. Random-window validation is a diagnostic of within-corpus discrimination, not an estimate of unseen-station performance~\cite{Roberts2017:SpatialCV, Ploton2020:SpatialValidation}. Confidence intervals should resample the grouping units defining the target population, and pooled metrics should be accompanied by domain-specific evaluations and areas of applicability~\cite{Meyer2021:AreaApplicability}.

Scale helps transfer only when it adds relevant support. More recordings from represented deployments can reduce within-domain estimation error, but they cannot substitute for independent station coverage. In the qualitative framework of Eq.~\eqref{eq:bound}, additional observations may reduce source error while leaving source--target divergence unchanged~\cite{Wang2023:DomainGenSurvey}. For cross-station inference, the number, diversity, and acquisition coverage of stations are more relevant measures of independent support than raw recording count alone. The present scale experiment is too limited to estimate how performance changes with additional independent stations.

The measurements motivate evaluating a \emph{spatially expert} model whose experts adapt to coherent acoustic domains~\cite{Dryden2022:SpatialMoE}. A candidate router could use measured environmental, propagation, and acquisition affinity rather than geographic distance alone, with the four regions serving as one testable partition. A shared backbone could allow data-poor stations to borrow strength, while regional or domain-specific adapters retain local structure. Contrastive or foundation representations are plausible candidate backbones~\cite{Acs2026:ContrastivePAM, Robinson2024:NatureLMaudio}. An area-of-applicability gate could abstain outside supported domains~\cite{Meyer2021:AreaApplicability}. These are hypotheses for future comparison, not remedies established by the present experiments, and station-grouped validation would remain necessary because any model can exploit acquisition and prior shortcuts.

Beyond modeling, vessel-detection technology has potential dual-use implications because methods that support conservation and marine management can also inform maritime security~\cite{Pallotta2013:AISAnomaly}. We study de-identified aggregate statistics, yet observed heterogeneity remains a governance concern because station- or acquisition-specific artifacts may fail when deployed elsewhere. Because label provenance depends on protected records, reproducibility relies on permitted derived tables and open code rather than release of source AIS records. Spatial sensing studies should treat acquisition configuration and station identity as potential confounders, quantify heterogeneity and areas of applicability, and compare pooled models with domain-aware alternatives under grouped validation.

\section{Conclusion}
\label{sec:conclusion}
This study examined dependence and transfer in a 908-thousand-recording underwater-acoustic corpus. AIS-contact prior varies by more than $200\times$, screening distributions require local interpretation, and station identifier is only weakly recoverable across seasons, at approximately 16\% balanced accuracy against a 5.6\% chance level. On the capped and retrospectively screened benchmark, random-window vessel detection exceeds unseen-station detection by a paired AUC difference of 0.049 (0.040 to 0.056), while station-bootstrap intervals represent uncertainty at the grouping level rather than treating recordings as independent. Training-station coverage shows a small, non-monotonic association with transfer in an exploratory analysis that is not a controlled learning curve.

Underwater Acoustic Big Data should therefore be treated as a geospatial domain-generalization problem, characterized for heterogeneity and validated by station identifier when unseen-station transfer is claimed. A spatial-expert architecture routed by acoustic and propagation affinity over a shared backbone remains a promising hypothesis rather than an established solution. Raw recording count should not be interpreted as effective independent sample size when observations share deployment and station context. Stronger transfer claims require broader independent spatial and acquisition support, along with models evaluated with respect to data origin.

\section*{Acknowledgements}
The authors thank JASCO Applied Sciences for the Passive Acoustic Monitoring data and Fisheries and Oceans Canada and the Bedford Institute of Oceanography for facilitating access to the Canadian Coast Guard AIS data used to construct the labels. This work was partially supported by the Natural Sciences and Engineering Research Council of Canada and the Conselho Nacional de Desenvolvimento Cient\'{i}fico e Tecnol\'{o}gico.

\section*{Data Availability}
The source acoustic recordings are available through NOAA's National Centers for Environmental Information under DOI 10.25921/sbnr-8505~\cite{NOAA2025:ESRFRawAudio}. The analysis repository at \url{https://github.com/gabrielspadon/scotian-shelf-labeling} contains the labeling and feature-extraction pipeline, the paper-analysis submodule, committed result tables, figure-generation code, manifests for station coordinates and hydrophone models, and a redistributable 908{,}072-recording feature table. Source Canadian Coast Guard AIS records cannot be redistributed. The public derived materials support reproduction of the downstream attribution, detection, robustness, feature-importance, and distance analyses, but not independent reconstruction of labels from the protected AIS archive.

\section*{Disclosure, Ethics, and Responsible AI Use}
This study uses pre-collected acoustic and vessel-traffic records, involves no human subjects, and does not intervene in marine operations. Sensitive data were handled in accordance with access agreements, and protected vessel-level records are not reported. The authors declare no competing interests. Generative AI technologies assisted with drafting and code scaffolding; the authors are responsible for verifying all claims, analyses, figures, and results before submission.

\bibliographystyle{IEEEtran}
\bibliography{references}

@IEEEtranBSTCTL{IEEEtran:BSTcontrol,
  CTLuse_forced_etal       = "yes",
  CTLmax_names_forced_etal = "3",
  CTLnames_show_etal       = "1"
}

@article{Tobler1970:FirstLaw,
  author  = {Tobler, Waldo R.},
  title   = {{A Computer Movie Simulating Urban Growth in the {Detroit} Region}},
  journal = {Econ. Geogr.},
  volume  = {46},
  pages   = {234--240},
  year    = {1970},
  doi     = {10.2307/143141}
}

@article{Legendre1993:SpatialAutocorr,
  author  = {Legendre, Pierre},
  title   = {{Spatial Autocorrelation: Trouble or New Paradigm?}},
  journal = {Ecology},
  volume  = {74},
  number  = {6},
  pages   = {1659--1673},
  year    = {1993},
  doi     = {10.2307/1939924}
}

@article{MorenoTorres2012:ShiftView,
  author  = {Moreno-Torres, Jose G. and Raeder, Troy and Alaiz-Rodr\'{i}guez, Roc\'{i}o and Chawla, Nitesh V. and Herrera, Francisco},
  title   = {{A Unifying View on Dataset Shift in Classification}},
  journal = {Pattern Recognit.},
  volume  = {45},
  number  = {1},
  pages   = {521--530},
  year    = {2012},
  doi     = {10.1016/j.patcog.2011.06.019}
}

@article{BenDavid2010:DomainTheory,
  author  = {Ben-David, Shai and Blitzer, John and Crammer, Koby and Kulesza, Alex and Pereira, Fernando and Vaughan, Jennifer Wortman},
  title   = {{A Theory of Learning from Different Domains}},
  journal = {Mach. Learn.},
  volume  = {79},
  number  = {1--2},
  pages   = {151--175},
  year    = {2010},
  doi     = {10.1007/s10994-009-5152-4}
}

@article{Wang2023:DomainGenSurvey,
  author  = {Wang, Jindong and Lan, Cuiling and Liu, Chang and Ouyang, Yidong and Qin, Tao and Lu, Wang and Chen, Yiqiang and Zeng, Wenjun and Yu, Philip S.},
  title   = {{Generalizing to Unseen Domains: A Survey on Domain Generalization}},
  journal = {IEEE Trans. Knowl. Data Eng.},
  volume  = {35},
  number  = {8},
  pages   = {8052--8072},
  year    = {2023},
  doi     = {10.1109/TKDE.2022.3178128}
}

@article{Geirhos2020:ShortcutLearning,
  author  = {Geirhos, Robert and Jacobsen, J{\"o}rn-Henrik and Michaelis, Claudio and Zemel, Richard and Brendel, Wieland and Bethge, Matthias and Wichmann, Felix A.},
  title   = {{Shortcut Learning in Deep Neural Networks}},
  journal = {Nat. Mach. Intell.},
  volume  = {2},
  number  = {11},
  pages   = {665--673},
  year    = {2020},
  doi     = {10.1038/s42256-020-00257-z}
}

@inproceedings{Beery2018:TerraIncognita,
  author    = {Beery, Sara and Van Horn, Grant and Perona, Pietro},
  title     = {{Recognition in Terra Incognita}},
  booktitle = {Proc. Eur. Conf. Comput. Vis. (ECCV)},
  pages     = {456--473},
  year      = {2018},
  doi       = {10.1007/978-3-030-01270-0_28}
}

@article{Frenay2014:LabelNoise,
  author  = {Fr\'{e}nay, Beno\^{i}t and Verleysen, Michel},
  title   = {{Classification in the Presence of Label Noise: A Survey}},
  journal = {IEEE Trans. Neural Netw. Learn. Syst.},
  volume  = {25},
  number  = {5},
  pages   = {845--869},
  year    = {2014},
  doi     = {10.1109/TNNLS.2013.2292894}
}

@article{Janowicz2020:GeoAI,
  author  = {Janowicz, Krzysztof and Gao, Song and McKenzie, Grant and Hu, Yingjie and Bhaduri, Budhendra},
  title   = {{{GeoAI}: Spatially Explicit Artificial Intelligence Techniques for Geographic Knowledge Discovery and Beyond}},
  journal = {Int. J. Geogr. Inf. Sci.},
  volume  = {34},
  number  = {4},
  pages   = {625--636},
  year    = {2020},
  doi     = {10.1080/13658816.2019.1684500}
}

@article{Reichstein2019:DLEarthSystem,
  author  = {Reichstein, Markus and Camps-Valls, Gustau and Stevens, Bjorn and Jung, Martin and Denzler, Joachim and Carvalhais, Nuno and Prabhat},
  title   = {{Deep Learning and Process Understanding for Data-Driven {Earth} System Science}},
  journal = {Nature},
  volume  = {566},
  number  = {7743},
  pages   = {195--204},
  year    = {2019},
  doi     = {10.1038/s41586-019-0912-1}
}

@article{Roberts2017:SpatialCV,
  author  = {Roberts, David R. and Bahn, Volker and Ciuti, Simone and Boyce, Mark S. and Elith, Jane and Guillera-Arroita, Gurutzeta and Hauenstein, Severin and Lahoz-Monfort, Jos\'{e} J. and Schr\"{o}der, Boris and Thuiller, Wilfried and Warton, David I. and Wintle, Brendan A. and Hartig, Florian and Dormann, Carsten F.},
  title   = {{Cross-Validation Strategies for Data with Temporal, Spatial, Hierarchical, or Phylogenetic Structure}},
  journal = {Ecography},
  volume  = {40},
  number  = {8},
  pages   = {913--929},
  year    = {2017},
  doi     = {10.1111/ecog.02881}
}

@article{Ploton2020:SpatialValidation,
  author  = {Ploton, Pierre and Mortier, Fr\'{e}d\'{e}ric and R\'{e}jou-M\'{e}chain, Maxime and Barbier, Nicolas and Picard, Nicolas and Rossi, Vivien and Dormann, Carsten and Cornu, Guillaume and Viennois, Ga\"{e}lle and Bayol, Nicolas and Lyapustin, Alexei and Gourlet-Fleury, Sylvie and P\'{e}lissier, Rapha\"{e}l},
  title   = {{Spatial Validation Reveals Poor Predictive Performance of Large-Scale Ecological Mapping Models}},
  journal = {Nat. Commun.},
  volume  = {11},
  number  = {1},
  pages   = {4540},
  year    = {2020},
  doi     = {10.1038/s41467-020-18321-y}
}

@article{Meyer2021:AreaApplicability,
  author  = {Meyer, Hanna and Pebesma, Edzer},
  title   = {{Predicting into Unknown Space? {Estimating} the Area of Applicability of Spatial Prediction Models}},
  journal = {Methods Ecol. Evol.},
  volume  = {12},
  number  = {9},
  pages   = {1620--1633},
  year    = {2021},
  doi     = {10.1111/2041-210X.13650}
}

@book{Urick1983:UnderwaterSound,
  author    = {Urick, Robert J.},
  title     = {{Principles of Underwater Sound}},
  edition   = {3rd},
  publisher = {McGraw-Hill},
  address   = {New York, NY, USA},
  year      = {1983}
}

@article{Wenz1962:AmbientNoise,
  author  = {Wenz, Gordon M.},
  title   = {{Acoustic Ambient Noise in the Ocean: Spectra and Sources}},
  journal = {J. Acoust. Soc. Am.},
  volume  = {34},
  number  = {12},
  pages   = {1936--1956},
  year    = {1962},
  doi     = {10.1121/1.1909155}
}

@article{Blondin2025:VesselStrikeRisk,
  author  = {Blondin, Hannah and Garrison, Lance P. and Adams, Jeffrey D. and Roberts, Jason J. and Good, Caroline P. and Gahm, Meghan P. and Lisi, Niki E. and Patterson, Eric M.},
  title   = {{Vessel Strike Encounter Risk Model Informs Mortality Risk for Endangered {North Atlantic} Right Whales Along the {United States} East Coast}},
  journal = {Sci. Rep.},
  volume  = {15},
  pages   = {736},
  year    = {2025},
  doi     = {10.1038/s41598-024-84886-z}
}

@article{Erbe2019:ShipNoiseReview,
  author  = {Erbe, Christine and Marley, Sarah A. and Schoeman, Renee P. and Smith, Joshua N. and Trigg, Leah E. and Embling, Clare B.},
  title   = {{The Effects of Ship Noise on Marine Mammals---A Review}},
  journal = {Front. Mar. Sci.},
  volume  = {6},
  pages   = {606},
  year    = {2019},
  doi     = {10.3389/fmars.2019.00606}
}

@article{Merchant2015:AcousticHabitats,
  author  = {Merchant, Nathan D. and Fristrup, Kurt M. and Johnson, Mark P. and Tyack, Peter L. and Witt, Matthew J. and Blondel, Philippe and Parks, Susan E.},
  title   = {{Measuring Acoustic Habitats}},
  journal = {Methods Ecol. Evol.},
  volume  = {6},
  number  = {3},
  pages   = {257--265},
  year    = {2015},
  doi     = {10.1111/2041-210X.12330}
}

@article{Stowell2022:Bioacoustics,
  author  = {Stowell, Dan},
  title   = {{Computational Bioacoustics with Deep Learning: A Review and Roadmap}},
  journal = {PeerJ},
  volume  = {10},
  pages   = {e13152},
  year    = {2022},
  doi     = {10.7717/peerj.13152}
}

@article{Schall2024:BaleenBenchmark,
  author  = {Schall, Elena and Kaya, Idil Ilgaz and Debusschere, Elisabeth and Devos, Paul and Parcerisas, Clea},
  title   = {{Deep Learning in Marine Bioacoustics: A Benchmark for Baleen Whale Detection}},
  journal = {Remote Sens. Ecol. Conserv.},
  volume  = {10},
  number  = {5},
  pages   = {642--654},
  year    = {2024},
  doi     = {10.1002/rse2.392}
}

@article{Acs2026:ContrastivePAM,
  author  = {Acs, Richard and Ibrahim, Ali and Zhuang, Hanqi and Ch\'{e}rubin, Laurent M.},
  title   = {{Contrastive Learning for Passive Acoustic Monitoring: A Framework for Sound Source Discovery and Cross-Site Comparison in Marine Soundscapes}},
  journal = {PLoS Comput. Biol.},
  volume  = {22},
  number  = {3},
  pages   = {e1014005},
  year    = {2026},
  doi     = {10.1371/journal.pcbi.1014005}
}

@article{Pallotta2013:AISAnomaly,
  author  = {Pallotta, Giuliana and Vespe, Michele and Bryan, Karna},
  title   = {{Vessel Pattern Knowledge Discovery from {AIS} Data: A Framework for Anomaly Detection and Route Prediction}},
  journal = {Entropy},
  volume  = {15},
  number  = {6},
  pages   = {2218--2245},
  year    = {2013},
  doi     = {10.3390/e15062218}
}

@article{Spadon2024:VesselForecast,
  author  = {Spadon, Gabriel and Kumar, Jay and Eden, Derek and van Berkel, Josh and Foster, Tom and Soares, Amilcar and Fablet, Ronan and Matwin, Stan and Pelot, Ronald},
  title   = {{Multi-Path Long-Term Vessel Trajectories Forecasting with Probabilistic Feature Fusion for Problem Shifting}},
  journal = {Ocean Eng.},
  volume  = {312},
  pages   = {119138},
  year    = {2024},
  doi     = {10.1016/j.oceaneng.2024.119138}
}

@article{Spadon2024:AISdb,
  author  = {Spadon, Gabriel and Kumar, Jay and Chen, Jinkun and Smith, Matthew and Hilliard, Casey and Vela, Sarah and Gehrmann, Romina and DiBacco, Claudio and Matwin, Stan and Pelot, Ronald},
  title   = {{Maritime Tracking Data Analysis and Integration with {AISdb}}},
  journal = {SoftwareX},
  volume  = {28},
  pages   = {101952},
  year    = {2024},
  doi     = {10.1016/j.softx.2024.101952}
}

@inproceedings{Liu2008:IsolationForest,
  author    = {Liu, Fei Tony and Ting, Kai Ming and Zhou, Zhi-Hua},
  title     = {{Isolation Forest}},
  booktitle = {Proc. IEEE Int. Conf. Data Mining (ICDM)},
  pages     = {413--422},
  year      = {2008},
  doi       = {10.1109/ICDM.2008.17}
}

@article{Welch1967:PSD,
  author  = {Welch, Peter D.},
  title   = {{The Use of Fast {Fourier} Transform for the Estimation of Power Spectra: A Method Based on Time Averaging over Short, Modified Periodograms}},
  journal = {IEEE Trans. Audio Electroacoust.},
  volume  = {15},
  number  = {2},
  pages   = {70--73},
  year    = {1967},
  doi     = {10.1109/TAU.1967.1161901}
}

@book{Bellman1961:Curse,
  author    = {Bellman, Richard E.},
  title     = {{Adaptive Control Processes: A Guided Tour}},
  publisher = {Princeton Univ. Press},
  address   = {Princeton, NJ, USA},
  year      = {1961}
}

@misc{Kaplan2020:ScalingLaws,
  author       = {Kaplan, Jared and McCandlish, Sam and Henighan, Tom and Brown, Tom B. and Chess, Benjamin and Child, Rewon and Gray, Scott and Radford, Alec and Wu, Jeffrey and Amodei, Dario},
  title        = {{Scaling Laws for Neural Language Models}},
  howpublished = {arXiv:2001.08361},
  year         = {2020},
  doi          = {10.48550/arXiv.2001.08361}
}

@techreport{Delarue2018:ESRFReport,
  author      = {Delarue, Julien J.-Y. and Kowarski, Katie A. and Maxner, Emily E. and MacDonnell, Jeff T. and Martin, S. Bruce},
  title       = {{Acoustic Monitoring Along {Canada's} East Coast: {August} 2015 to {July} 2017}},
  institution = {Environ. Stud. Res. Funds},
  number      = {215},
  address     = {Dartmouth, NS, Canada},
  year        = {2018},
  note        = {{JASCO} Document 01279}
}

@article{Kowarski2021:BigDataPAM,
  author  = {Kowarski, Katie A. and Moors-Murphy, Hilary},
  title   = {{A Review of Big Data Analysis Methods for Baleen Whale Passive Acoustic Monitoring}},
  journal = {Mar. Mammal Sci.},
  volume  = {37},
  number  = {2},
  pages   = {652--673},
  year    = {2021},
  doi     = {10.1111/mms.12758}
}

@article{Martin2021:MillidecadeSpectra,
  author  = {Martin, S. Bruce and Gaudet, Briand J. and Klinck, Holger and Dugan, Peter J. and Miksis-Olds, Jennifer L. and Mellinger, David K. and Mann, David A. and Boebel, Olaf and Wilson, Colleen C. and Ponirakis, Dimitri W. and Moors-Murphy, Hilary},
  title   = {{Hybrid Millidecade Spectra: A Practical Format for Exchange of Long-Term Ambient Sound Data}},
  journal = {JASA Express Lett.},
  volume  = {1},
  number  = {1},
  pages   = {011203},
  year    = {2021},
  doi     = {10.1121/10.0003324}
}

@inproceedings{Dryden2022:SpatialMoE,
  author    = {Dryden, Nikoli and Hoefler, Torsten},
  title     = {{Spatial Mixture-of-Experts}},
  booktitle = {Proc. Adv. Neural Inf. Process. Syst. (NeurIPS)},
  volume    = {35},
  year      = {2022}
}

@misc{NOAA2025:ESRFRawAudio,
  author       = {{NOAA National Centers for Environmental Information}},
  title        = {{{ESRF Atlantic Canada} Passive Acoustic Monitoring 2015--2017 Raw Audio}},
  howpublished = {NOAA National Centers for Environmental Information},
  year         = {2025},
  doi          = {10.25921/sbnr-8505},
  note         = {Dataset, accessed 2026-07-15}
}

@misc{Robinson2024:NatureLMaudio,
  author       = {Robinson, David and Miron, Marius and Hagiwara, Masato and Weck, Benno and Keen, Sara and Alizadeh, Milad and Narula, Gagan and Geist, Matthieu and Pietquin, Olivier},
  title        = {{{NatureLM-audio}: An Audio-Language Foundation Model for Bioacoustics}},
  howpublished = {arXiv:2411.07186},
  year         = {2024},
  doi          = {10.48550/arXiv.2411.07186}
}

\end{document}